\documentclass[conference]{IEEEtran}
\IEEEoverridecommandlockouts

\usepackage{cite}
\usepackage{amsmath,amssymb,amsfonts}
\usepackage{algorithm}
\usepackage{algpseudocode}
\usepackage{graphicx}
\usepackage{textcomp}
\usepackage{verbatim}
\usepackage{xcolor}
\usepackage{url}
\usepackage{subfigure}
\def\BibTeX{{\rm B\kern-.05em{\sc i\kern-.025em b}\kern-.08em
    T\kern-.1667em\lower.7ex\hbox{E}\kern-.125emX}}

\begin{document}

\title{MLRS-PDS: A Meta-learning recommendation of dynamic ensemble selection pipelines\\
}

\author{\IEEEauthorblockN{Hesam Jalalian and Rafael M. O. Cruz}
\IEEEauthorblockA{École de Technologie Supérieure, Université du Québec, Montréal (QC), Canada\\
Email: hesam.jalalian.1@ens.etsmtl.ca, rafael.menelau-cruz@etsmtl.ca}}

\maketitle
\pagestyle{empty}

\begin{abstract}

Dynamic Selection (DS), where base classifiers are chosen from a classifier's pool for each new instance at test time, has shown to be highly effective in pattern recognition.
However, instability and redundancy in the classifier pools can impede computational efficiency and accuracy in dynamic ensemble selection. This paper introduces a meta-learning recommendation system (MLRS) to recommend the optimal pool generation scheme for DES methods tailored to individual datasets. The system employs a meta-model built from dataset meta-features to predict the most suitable pool generation scheme and DES method for a given dataset. Through an extensive experimental study encompassing 288 datasets, we demonstrate that this meta-learning recommendation system outperforms traditional fixed pool or DES method selection strategies, highlighting the efficacy of a meta-learning approach in refining DES method selection. The source code, datasets, and supplementary results can be found in this project's GitHub repository: \url{https://github.com/Menelau/MLRS-PDS}.

\end{abstract}


\section{Introduction}

Automated decision-making frequently utilizes Multiple Classifier Systems (MCS), wherein individual classifiers' collective predictions enhance overall prediction accuracy \cite{polikar2006ensemble}. The process of MCS unfolds in three key stages: generation, involving the formation of a pool of classifiers; selection, which entails either a static or dynamic choice among these classifiers; and aggregation, where the outputs from the selected experts are combined to formulate the final decision \cite{cruz2018dynamic}.

Dynamic selection techniques (DS) within Multiple Classifier Systems (MCS) are characterized by their approach of selecting classifiers during test time, considering the characteristics of each specific instance to improve prediction accuracy. Such methods are based on the assumption that each base classifier is an expert in a different local region of the feature space, and only the classifiers that are experts in the local region where the test instance is located should be used to predict its label~\cite{cruz2018dynamic}. They operate under the locality assumption, suggesting that similar instances should share the same set of expert classifiers~\cite{souza2024dynamic}. DS has become increasingly prominent in various pattern recognition contexts. These include imbalanced learning \cite{roy2018study}, handling noisy data \cite{walmsley2022investigation}, and adapting to concept drift \cite{almeida2018adapting}, showing its broad impact.

For optimal performance, DS requires a well-suited pool of classifiers that effectively covers the entire feature space and that for each instance there exists, enabling the DS method to accurately identify the best classifiers for any given instance \cite{cruz2019fire}. However, the literature offers limited guidance on selecting or training these classifier pools specifically for DS algorithms, with only a few studies addressing this gap \cite{souza2019online, monteiro2023exploring}. While several DS studies have utilized conventional pool generation schemes such as Bagging \cite{breiman1996bagging}, Boosting \cite{freund1997decision} and Random Forests \cite{rodrigues2023security}, these methods were originally designed for static combinations and may not fully align with the local assumptions of DS models, potentially leading to gaps in expert coverage across the feature space \cite{oliveira2017online}. Current research predominantly focuses on refining competence level estimation, ensemble selection, or defining regions of competence to boost performance. This trend often suggests an assumption that classifier pools generated by classical ensemble techniques like Bagging or Boosting are universally effective for all classification challenges, ignoring the crucial role played by the composition of the classifier pool.

We hypothesize that the pool generation scheme plays a crucial role in the performance of DS methods and should not be neglected. Furthermore, the choice of pool generation scheme must consider the dataset's properties and the dynamic selection technique used for classification since distinct DS techniques are based on different local assumptions. Specifically, this research addresses the following question: \emph{What are the consequences of neglecting the selection of either the DS method or the pool generation scheme on the performance of the dynamic selection pipeline? Furthermore, is it possible to automate this selection process?}

In this paper, we propose a novel meta-learning recommendation system (MLRS) designed to enhance the performance of Dynamic Selection (DS) methods while reducing the complexity of finding optimal solutions. It employs meta-learning to extract dataset characteristics (i.e., meta-features) alongside the performance evaluation data from various classifier pools used as input for DS algorithms to learn how to recommend the best pool and/or DS algorithm given a new dataset.

We propose three distinct MLRS variants operating in three scenarios: 1) MLRS-P that recommends the best pool generation scheme based on the problem's characteristics and a user-specified DS model; 2) MLRS-DS that recommends the most effective DS method for a given problem, conditioned on a predefined existing pool specified; and 3) MLRS-PDS that automatically recommend the optimal (DS, Pool) pair solely based on the problem characteristics. MLRS-PDS first identifies the most suitable pool of classifiers according to the dataset characteristics. Then, it uses the information from the selected pool to condition the more suitable DS model recommendation. Thereby offering a fully automated meta-learning recommendation model for dynamic selection pipelines without requiring expert intervention.

To assess the efficacy of our proposed MLRS, we carried out extensive experiments using 288 datasets with varied complexity levels. The results consistently showed that relying on a fixed combination leads to sub-optimal results. Also, one needs to take into account the dependencies between data characteristics, the pool generation scheme, and the DS model, as the best choices can significantly change according to these parameters. In addition, our MLRS, in all its formulations, can recommend optimal solutions, being much more efficient than existing baselines.

The contributions of this work are summarized as follows:

\begin{itemize}
    \item It highlights the limitations of relying on either a fixed pool generation scheme or a fixed DS method, emphasizing the need to carefully optimize these steps.

    \item We propose a Meta-Learning Recommendation System (MLRS) for various use cases: recommending the most suitable pool generation scheme and DS method based on the unique characteristics of each dataset.

    \item We demonstrate that our proposed MLRS-PDS, which recommends both the pool generation scheme and the DS method simultaneously, leads to more optimal solutions than fixing either the pool or the DS method.

    \item An extensive analysis was conducted across 288 datasets with varying levels of complexity, revealing the advantages of our approach over traditional fixed pool and DS method selection strategies.
    \end{itemize}

\section{Related work}
\label{sec:related}

\noindent \textbf{Pool generation scheme.} The performance of DS methods hinges on a proper pool of classifiers, ensuring a diverse and complementary set of models is available that covers the whole feature space~\cite{souza2017characterization}. Pool generation can be broadly categorized into global and local perspectives. The Global pool generation scheme employs techniques that take a broad view of the problem and try to generate classifiers that model the whole data distribution. They initially proposed for static selection methods such as Bagging~\cite{breiman1996bagging}, Boosting~\cite{freund1997decision}, and Random Forests~\cite{breiman2001random}. In contrast, local pool generation schemes involve techniques that explicitly train classifiers that focus on modeling local regions of the feature space. Thus, methods with expertise in distinct feature space regions are obtained. These include Forest of Local Trees (FLT)\cite{armano2018building} and the Locally Independent Training (LIT) technique\cite{ross2020ensembles}. While most DS research favors global schemes, local perspective utilization is less common.

Monteiro et al. \cite{monteiro2023exploring} introduced a pool generation method focusing on diversity from data complexity and classifier decisions. This approach assesses the variability of complexity measures and employs an evolutionary algorithm to optimize complexity and decision diversity. The method trains classifiers on subsets of varying complexity and aims to produce classifiers with diverse error types. This global perspective pool generation method positively impacts DS methods by generating more diverse local experts through subsets of varying complexities. However, the framework lacks an automated process for selecting a base model suitable for the specific dataset. Moreover, the generated pool is not optimized for a particular DS method.

An online pool generation method \cite{souza2019online} creates local perspective pools for challenging feature space regions, employing specialized classifiers for instances prone to misclassification. This Local Pool (LP) approach uses Dynamic Classifier Selection (DCS) techniques to select competent classifiers for each instance located in class overlap regions. If a query instance falls in a complex region, an LP is dynamically generated for labeling; otherwise, a simple nearest neighbors rule is applied. These approaches, however, do not explore meta-learning, with the former focusing on a global pool generation perspective and the latter focusing on online learning and limited to DCS models.

\noindent \textbf{Meta-learning.} Conventional approaches for the algorithm selection problem~\cite{rice1976algorithm} rely on extensive expert knowledge and trial and error. However, they are extremely limited as trial and error is time-consuming and computationally expensive, making it impossible to cover all possible algorithm combinations. This problem led to a growing interest in machine learning systems that automate algorithm selection to address these challenges. One such approach is meta-learning-based algorithm recommendation~\cite{brazdil2022metalearning}, which treats algorithm selection as a typical learning problem. In meta-learning, dataset characteristics (meta-features) are the independent variables, and the target variable corresponds to the estimation of algorithm performance. This approach has found success in various domains, including classification \cite{garcia2018classifier}, clustering \cite{pimentel2019new}, and regression \cite{aguiar2022using}. By automating algorithm selection, these systems significantly reduce the computational cost required to tune solutions and empower non-experts to apply machine learning more independently \cite{khan2020literature}.

\begin{figure*}
	\centering
		\includegraphics[width=0.9\textwidth]{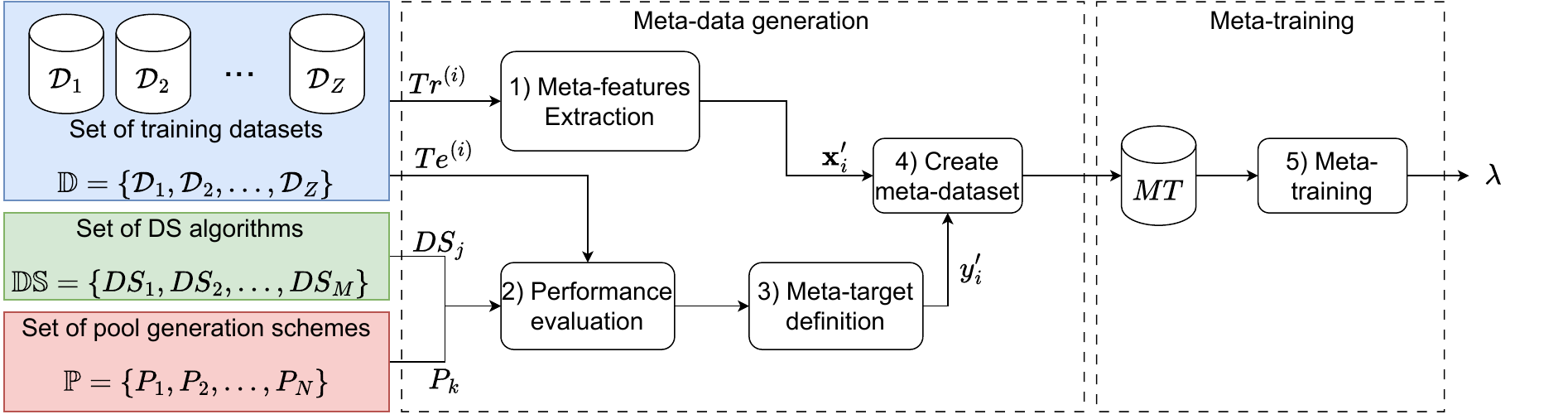}
	\caption{Overview of the meta-training process. In the first step, the meta-features, $mf$, are extracted from the training datasets to generate its representation $x'_{i}$. In step 2, the set of pools and DS methods are evaluated. Then, based on the highest accuracy, the meta-target, $y'$, is defined (step 3). In step 4, the meta-dataset, $MT$, is constructed, and then it is used to train a meta-model, $\lambda$ (Step 5)}\label{mtl_train} 
 \vspace{-1em}
\end{figure*}

Several works in meta-learning have been conducted recently, focusing on recommending different aspects of a machine-learning pipeline. In~\cite{gomes2012combining}, the authors proposed a meta-learning framework to recommend the hyperparameters of an SVM classifier. Garcia et al.\cite{garcia2018classifier} explored the use of complexity measures in meta-learning to differentiate classifier performance effectively. Their study demonstrated that it is possible to accurately predict expected classifier performances using meta-regressors and use such information to recommend the most appropriate models. Focusing on the preprocessing steps of an ML solution, Amorim et al.~\cite{de2024meta} proposed a meta-learning framework called Meta-Scaler to recommend the most appropriate scaling-transform technique according to the dataset characteristics and the classifier specified by the user. Building on this idea of meta-learning in the context of ensemble models, Pinto et al.\cite{pinto2017autobagging} proposed an automated bagging system using a meta-learning-based ranking approach learned from metadata. However, their focus was primarily on using bagging as a global perspective pool generation scheme without delving into the potential of local perspective pool generation schemes.

Regarding meta-learning recommendations for DS, the only work that can be found is based on meta-regression models for predicting the number of estimators required when applying a DS model~\cite{roy2016meta, roy2016metab}, demonstrating that often the best classification performance can be obtained by generating different pool sizes. However, this method is based on a fixed Bagging technique as a pool generation scheme and can only predict the pool size, thus leading to sub-optimal results. In contrast, our proposal recommends the pool generation algorithm instead of the pool size since the pool size does not significantly impact performance~\cite{cruz2015deep}. It also performs a chained recommendation model in which the system can recommend the pool generation algorithm and DS model together to obtain a final DS pipeline, differing from most current works in meta-learning that focus on predicting a single pipeline step.

\section{The Proposed Multi-label meta-learning recommendation (MLRS)}
\label{sec:proposed}

Before delving into the details of the proposed meta-learning recommendation system for DS methods, it is essential to define the basic concepts and mathematical notation used in this study. A dataset is represented by $\mathcal{D}$, where each dataset is a combination of a training partition $\mathcal{T}_r$ and a test partition $\mathcal{T}_e$. A set of datasets is symbolized by $\mathbb{D} = \{ \mathcal{D}_{1}, \ldots, \mathcal{D}_{Z} \}$, where $Z$ represents the number of datasets. The meta-feature vector (i.e., dataset characteristics) extracted from a dataset is denoted by $\mathbf{x'} = \{ mf_{1}, \ldots, mf_{d} \}$ where each $mf$ represent a meta-feature extracted from a dataset. The meta-target is indicated by $y'$. Finally, the meta-dataset is defined as $MT = \{ (\mathbf{x}'_1, y'_1), \ldots, (\mathbf{x}'_Z, y_Z') \}$, where each tuple $(\mathbf{x'_i}, y'_i)$ corresponding to the meta-features and meta-target extracted from a training dataset $\mathcal{D}_i$.

\subsection{MLRS Training process}

The meta-training stage is a crucial step within the meta-learning framework. Its objective is to create a meta-model, $\lambda$, that learns the relationship between the characteristics of a dataset and the performance of multiple models evaluated over it (Figure~\ref{mtl_train}. The training phase of the meta-learning recommendation system is detailed in Algorithm \ref{alg:mltrain}. The algorithm takes inputs: the set of datasets, $\mathbb{D}$, a set of pool generation scheme $\mathbb{P}$, and a set of DS algorithms $\mathbb{DS}$.

\begin{algorithm}
  \caption{MLRS training phase}
  \label{alg:mltrain}
  \begin{algorithmic}[1]
    \Require A set of training datasets, $\mathbb{D}$, a set of DS methods, $\mathbb{DS}$, a set of pools, $\mathbb{P}$
    \Ensure Meta-model, $\lambda$
    \State Initialize: $MT = \emptyset$
    \For{each $\mathcal{D}_i$ in $\mathbb{D}$}
      \State Extract meta-feature vector, $\mathbf{x}'_{i}$, from the dataset $\mathcal{T}_r^{(i)}$ 
      \For{each $DS_j \in \mathbb{DS}$}
        \For{each $P_k \in \mathbb{P}$}
          \State Assess the performance of $DS_j$ using $P_k$
        \EndFor
      \EndFor
      \State Define the configuration with the highest performance as $y'_{i}$
      \State $MT = MT \cup (\mathbf{x}'_{i}, y'_{i})$
    \EndFor
    \State Train the meta-model, $\lambda$, on the meta-dataset, $MT$
    \State \textbf{Return} $\lambda$
  \end{algorithmic}
\end{algorithm}

The algorithm then iterates over each dataset $\mathcal{D}_i \in \mathbb{D}$. For each iteration, the algorithm extracts the meta-features vector, $\mathbf{x}_{i}'$, capturing the characteristics of the training partition of the dataset $\mathcal{D}_i$ denoted by $\mathcal{T}_r^{(i)}$ (Section~\ref{sec:metafeatures}). Then, the combinations of pool and DS methods are assessed using its test partition denoted by $\mathcal{T}_e^{(i)}$. The configuration that obtains the highest performance is used to define the meta-target $y_i'$.

Then, a meta-dataset, $MT$, is constructed using the meta-feature vector, $\mathbf{x}_{i}'$, and the meta-target, $y_i'$, extracted from all training datasets. Subsequently, a meta-model $\lambda$ is trained using the meta-dataset, $MT$, to learn the mapping between the dataset characteristics and which models are more likely to obtain higher performance. The MLRS allows us to investigate three possible recommendation scenarios, each one generating a different meta-learning model $\lambda$ for the given task:

\begin{itemize}
    \item \textbf{Scenario I, MLRS-P:} The system is trained to recommend an optimal pool generation scheme conditioned to a specific DS method. In this meta-learning scenario, the meta-target $y' = y'_{pool}$ is the pool generation scheme that achieved the highest classification performance when used with the predefined DS method specified by the user. In this case, $\mathbb{DS}$ consists of a single model, that is, the predefined DS, and the meta-classifier $\lambda_{pool}$ is specifically trained to make recommendations for it.

    \item \textbf{Scenario II, MLRS-DS:} The system is trained to recommend an optimal DS method given a specific pool generation scheme. Here, the meta-target $y' = y'_{DS}$ is the DS method that exhibited the highest classification performance with the predetermined pool generation scheme. In this scenario, $\mathbb{P}$ consists of a single pool generation scheme, and the meta-classifier $\lambda_{DS}$ is specifically trained to make recommendations for it.

    \item \textbf{Scenario III, MLRS-PDS:} The system is trained to recommend an optimal pair (Pool, DS) based solely on the problem characteristics. The meta-target in this context is the tuple $(y'_{pool}, y'_{DS})$ that delivered the best performance compared to other configurations, representing a multi-label meta-learning recommendation problem. A classifier chain~\cite{read2011classifier} is employed to train the meta-classifier for this task, with the prediction order being recommending pool first, then recommending DS method conditioned on the first recommendation. As this scenario is fully automated and considers all possible configurations, both $\mathbb{P}$ and $\mathbb{DS}$ consist of multiple elements to generate the meta-training data $MT$.

\end{itemize}

\subsubsection{Meta-features}
\label{sec:metafeatures}
A crucial step in the meta-training stage is the extraction of meta-features, denoted as $mf$, from a collection of datasets, represented by $\mathbb{D}$. These meta-features serve as descriptors characterizing each dataset. In our study, we use a comprehensive set of 129 meta-features to capture the essential characteristics of the dataset, as suggested by Rivolli et al. \cite{rivolli2022meta}. These categories include Statistical, Information-theoretic, Model-based, Relative Landmarking \cite{reif2014automatic, smith2009cross, rivolli2022meta}, Clustering-based \cite{pimentel2019new}, Concept \cite{rivolli2022meta}, Itemset \cite{song2012automatic}, and Complexity \cite{ho2002complexity}. These meta-features were computed using the PyMFE library \cite{alcobacca2020mfe}, version 0.4.2. A complete list of these meta-features, their respective groups, and brief descriptions can be found in the supplementary material in the project's GitHub repository~\footnote{\url{https://github.com/Menelau/MLRS-PDS}}. Moreover, a comprehensive review of these meta-features is available in \cite{alcobacca2020mfe}.

\subsection{Meta-learning recommendation}

\begin{figure*}
\centering
\includegraphics[width=0.6\textwidth]{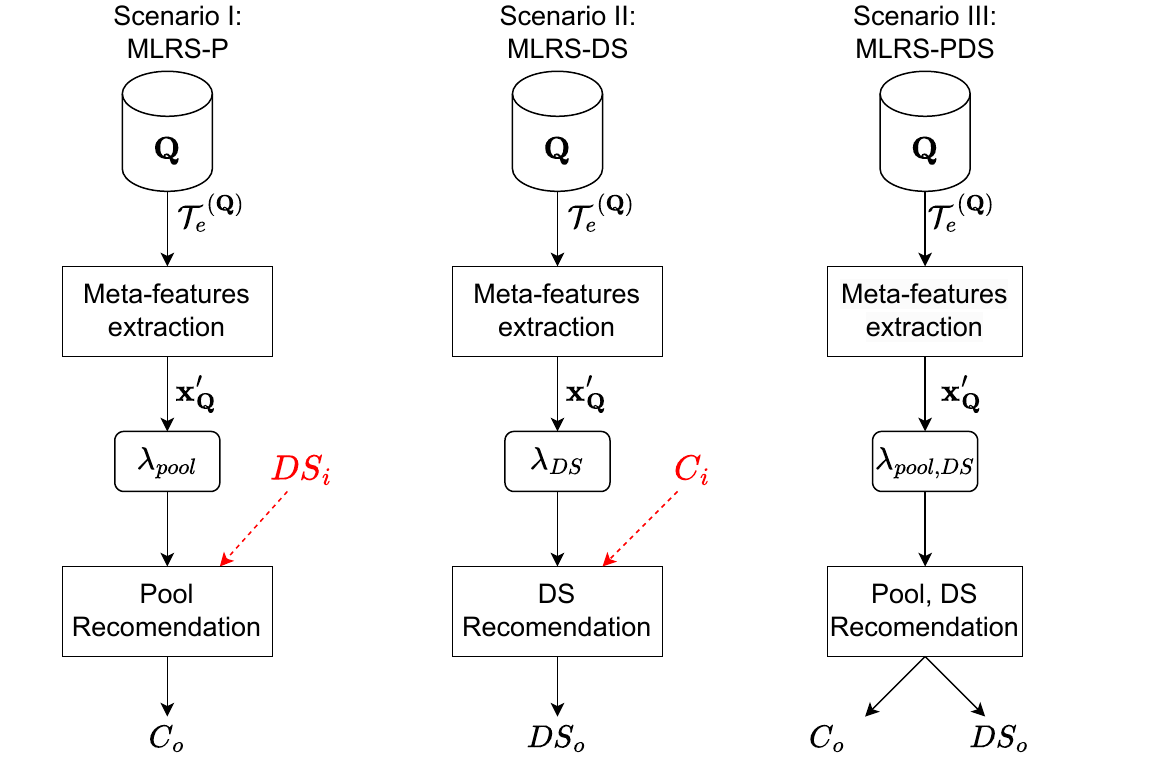}
\caption{The meta-learning recommendation process for the three distinct scenarios. The red arrow indicates the inputs (choices) provided by the user. In Scenario I, a pool generation scheme is recommended based on the dataset characteristics, conditional on the DS model specified by the user. Scenario II recommends a DS method based on the dataset characteristics and the pre-selected pool generation scheme. Scenario III recommends the best pair of (Pool, DS) without requiring user input. It is crucial to note that only the training set partition of the new query dataset $\mathbf{Q}$ is used for extracting meta-features, thereby preventing any data leakage from the test data.}
\label{mtl_gener}
\vspace{-1em}
\end{figure*}

Figure \ref{mtl_gener} depicts the recommendation phase of MLRS. It solely involves extracting the meta-feature representation of a query dataset and applying the meta-learner to predict the most suitable models. Consequently, our proposed system eliminates the need for extensive model selection evaluations, like grid search procedures, thereby significantly reducing computational demands. This process begins by considering a query dataset $\mathbf{Q}$, from which we extract its meta-feature representations, $\mathbf{x}'_{\mathbf{Q}}$ based on its available training set partition $\mathcal{T}_{r}^{\mathbf{Q}}$. This step is crucial for avoiding bias, specifically data leakage from the test distribution, in our meta-learning recommendation procedure. Following this, we utilize our trained meta-model, $\lambda$, to make dataset-specific recommendations. These recommendations correspond to one of the three recommendation scenarios:

\begin{itemize}
\item \textbf{Scenario I, MLRS-P:} The meta-learning recommendation system proposes a pool generation scheme while the DS method is fixed. It uses the meta-classifier $\lambda_{Pool}$ trained explicitly for this task. This scenario is applicable when a specific DS method must be used, and an optimal pool generation scheme, $P_o$, is required.

\item \textbf{Scenario II, MLRS-DS:} In this scenario, the meta-learning recommendation system suggests a DS method that complements a predetermined or pre-trained pool generation scheme. It employs the meta-classifier $\lambda_{DS}$, which was trained to recommend the best DS method for this predefined pool. This approach is beneficial when a particular pool generation scheme needs to be used, requiring identifying the most suitable DS method ($DS_o$) to obtain the best accuracy possible from this pool.

\item \textbf{Scenario III, MLRS-PDS:} This scenario involves the system recommending a pool and a DS model solely based on the dataset's characteristics. MLRS-PDS initially employs a classifier chain in a chained recommendation model to predict the optimal pool generation scheme using the meta-features. Subsequently, the selected pool's choice is incorporated as an input feature to aid in recommending the most suitable DS model contingent on it. This process automates the identification of the optimal Pool and DS pair ($P_o$, $DS_o$), effectively modeling the interdependencies between these two design choices and leading to an entire DS pipeline.

\end{itemize}

After executing the meta-learning recommendation process, the resultant tuple (Pool, DS) is employed to construct the DS pipeline. The selected pool generation scheme and the DS algorithm are trained using $\mathbf{Q}$ training partition ($\mathcal{T}_{r}^{(\mathbf{Q})}$). They are then applied in the generalization phase over the test dataset $\mathcal{T}_{e}^{(\mathbf{Q})}$ for labeling its instances.

\section{Experimental setup}
\label{sec:experiments}
\subsection{Pool generation schemes}

We explore seven distinct pool generation schemes, each carefully selected to represent pools of diverse natures. We consider the Bagging technique with linear Perceptrons (BP) and decision trees (BDT) as base models for global pool generation. Additionally, we employ Perceptrons and Decision Trees in conjunction with Adaboost, resulting in the BSP and BSDT pools, respectively. The Random Forest model (RF) was also included as another widely recognized global pool generation scheme in DS literature~\cite{rodrigues2023security,islam2019semantics}. These models are frequently utilized in various studies as pool generation methods, as evidenced by multiple references~\cite{souza2024dynamic, souza2023olp++, cruz2017meta, monteiro2023exploring, garcia2018dynamic, islam2019semantics, elmi2021multi, rodrigues2023security, davtalab2024scalable}. Thus encompassing a range of pool generation algorithms prominently featured in recent DS publications.

As local pool generation, we consider two methods: Forest of Local Trees (FLT)~\cite{armano2018building} and Locally Independent Training (LIT)~\cite{ross2020ensembles}. Notably, DS methods have not previously used local pool generation schemes in their conception. Given that DS methods rely on the local expert assumption, we hypothesize that such methods could present a viable alternative as pool generation algorithms for DS, potentially enhancing their classification performance. Including these methods thus broadens the scope of our MLRS, providing a more comprehensive examination of pool generation possibilities. The pool of classifiers in all simulations consists of $100$ base classifiers, following the recommendation in~\cite{cruz2015deep}. This uniform pool size ensures a fair comparison across all pool generation schemes.

\subsection{DS Techniques}

In this research, we utilized seven dynamic selection algorithms from the DESlib version 0.3.5 DS library \cite{cruz2020deslib}, incorporating a mix of two Dynamic Classifier Selection (DCS) and five Dynamic Ensemble Selection (DES) methods to ensure diverse and meaningful results. The DCS methods included Overall Local Accuracy (OLA) \cite{woods1997combination} and Modified Local Accuracy (MLA) \cite{smits2002multiple}, while the DES methods encompassed KNORA-E \cite{ko2008dynamic}, KNORA-U \cite{ko2008dynamic}, Meta-learning for dynamic ensemble selection (META-DES) \cite{cruz2015meta}, Multiclass Imbalance (DES-MI) \cite{garcia2018dynamic}, and Dynamic Ensemble Selection performance (DES-P) \cite{woloszynski2012measure}. All these techniques rely on the K-Nearest Neighbors (KNN) method with a region of competence size (K) set to 7, as per \cite{cruz2015meta}. Specifically for the META-DES algorithm, we use the configuration proposed in~\cite{cruz2017meta} comprising of the Naive Bayes algorithm for the meta-level and using a total of five output profiles.

\subsection{Datasets}
\label{sec:dataset}
Relying on a standard test bed without considering the diversity of dataset characteristics can lead to incomplete evaluations of learning algorithms. This limitation is particularly pronounced in meta-learning, where constructing a robust meta-classifier and comprehensively evaluating the meta-learning model require a diverse dataset collection. Hence, in this work, we considered the datasets from the Landscape Contest at ICPR 2010 \cite{macia2010landscape}, which consists of 301 datasets specially crafted to cover the space of dataset complexity~\cite{ho2002data}. Therefore, offering a well-rounded and comprehensive basis for addressing our research questions. 

In this study, 13 datasets were excluded from the original 301 due to issues with the AdaBoost algorithm using a Perceptron as the base estimator. These datasets, specifically numbered 216, 219, 220, 221, 252, 253, 254, 255, 257, 258, 260, 262, and 263, were problematic because AdaBoost, requiring a diverse set of classifiers, could only generate one classifier for each, failing to generate multiple models. .

\subsection{Experimental setup}
For this experiment, we employed the leave-one-dataset-out (LODO) procedure, where, at each iteration, one dataset ($\mathbf{Q}$) is left for testing, and the remaining ones are used as training datasets ($\mathbb{D}$). In other words, for each simulation, 287 datasets were considered for training the meta-learning framework, while one was considered as the test dataset $\mathbf{Q}$ used to evaluate its generalization performance. We divided each dataset into 75\% of the data used for training ($\mathcal{T}_r$) and the remaining 25\% for the data used for testing ($\mathcal{T}_e$) using a stratified holdout split. Datasets were normalized using the Z-score normalization, also known as Standard Scaler~\cite{de2023choice}.

\subsection{Meta-learner definition}
For each recommendation scenario, three algorithms, including Random Forests (RF), K Nearest Neighbors (KNN), and Support Vector Machine (SVM), were initially evaluated as the meta-modal. Their choices were based on previous meta-learning studies \cite{khan2020literature, rivolli2022meta}. The hyperparameter tuning procedure was conducted through a 10-fold cross-validation with a grid search. For K Nearest Neighbors, values of K ranging from 2 to 7 were evaluated, while for Random Forests, Max Depth values of 2 to 5 were tried. The hyperparameter chosen for the Support Vector Machine was gamma equal and cost varying between 0.01, 0.1, and 1. The best meta-model configuration found were RF with a Max\_Depth = 5 and 100 trees for Scenario I ($\lambda_{Pool})$ and KNN with K = 2 and using the Euclidean distance for Scenarios II and III ($\lambda_{DS}$ and $\lambda_{Pool, DS})$. 

\section{Results}
\label{sec:results}

This experimental study primarily aims to evaluate the performance of the three variants of our meta-learning recommendation system (MLRS) in recommending the best pool generation scheme, the best DS algorithm, and the entire pipeline. It is essential to highlight that this is the first work proposing automated algorithm selection for this task. Hence, we use established baselines from the Meta-learning literature, specifically a model that always recommends the majority class, i.e., the technique with the highest number of wins, for the entire testbed (Majority), and the average between all possible configurations which are common approaches to demonstrate the need for a recommendation system in meta-learning~\cite{brazdil2022metalearning, aguiar2022using}. We also present a statistical comparison between our meta-learning approach against all possible configurations for the pool and DS algorithm (a total of 49 configurations) in the supplementary material. 

Due to the vast amount of datasets and techniques involved ($7$ pool generation schemes $\times$ $7$ DS methods), we only present a synthesis of the results in the following sections. Classification accuracy of the corresponding recommended configuration for the base-level performance per dataset is detailed as supplementary material in the project's GitHub repository.

\subsection{Scenario I: meta-learning for recommending the best pool generation scheme}

In the first scenario, the MLRS-P recommends a pool generation scheme while the DS method is fixed. This recommendation scenario is applicable when a specific DS method should be used, requiring an optimized pool generation scheme for a given query dataset, $\mathbf{Q}$. This formulation has use cases, such as when performing a fair comparison between DS algorithms so that each one is optimized before evaluation or when a particular DS model needs to be used due to other constraints.

Table \ref{result1} compares the performance of DS methods employing a pool generation scheme recommended by MLRS-P against baseline approaches. Each row in the table pits MLRS-P against the Majority baseline — the pool generation scheme achieving the highest number of wins for the respective DS method — as well as against the average result across all possible combinations. The number of datasets for which the corresponding technique recommended the optimal method is indicated in parentheses.

\begin{figure}
  \centering
   \subfigure[]{\includegraphics[width=0.39\textwidth]{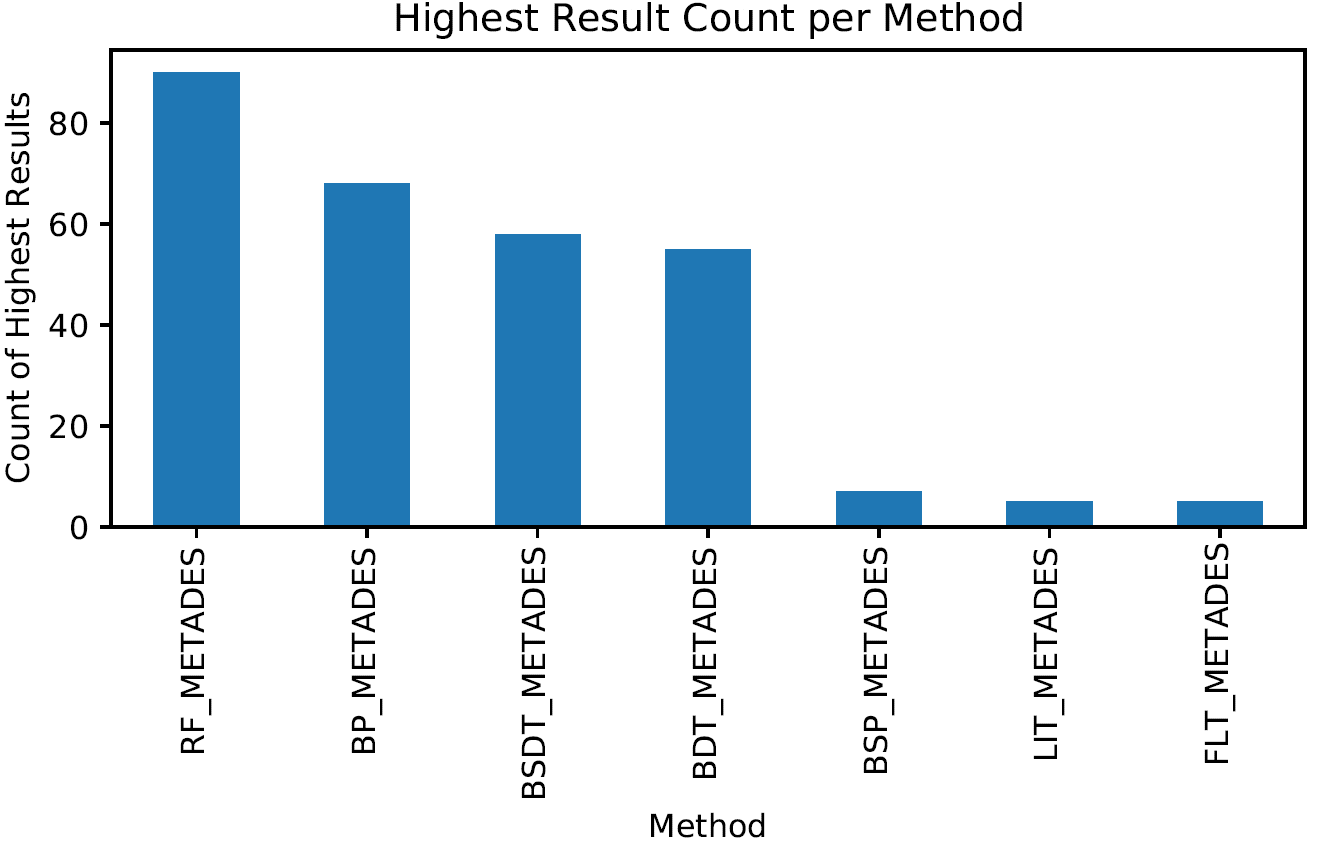}}
  \subfigure[]{\includegraphics[width=0.39\textwidth]{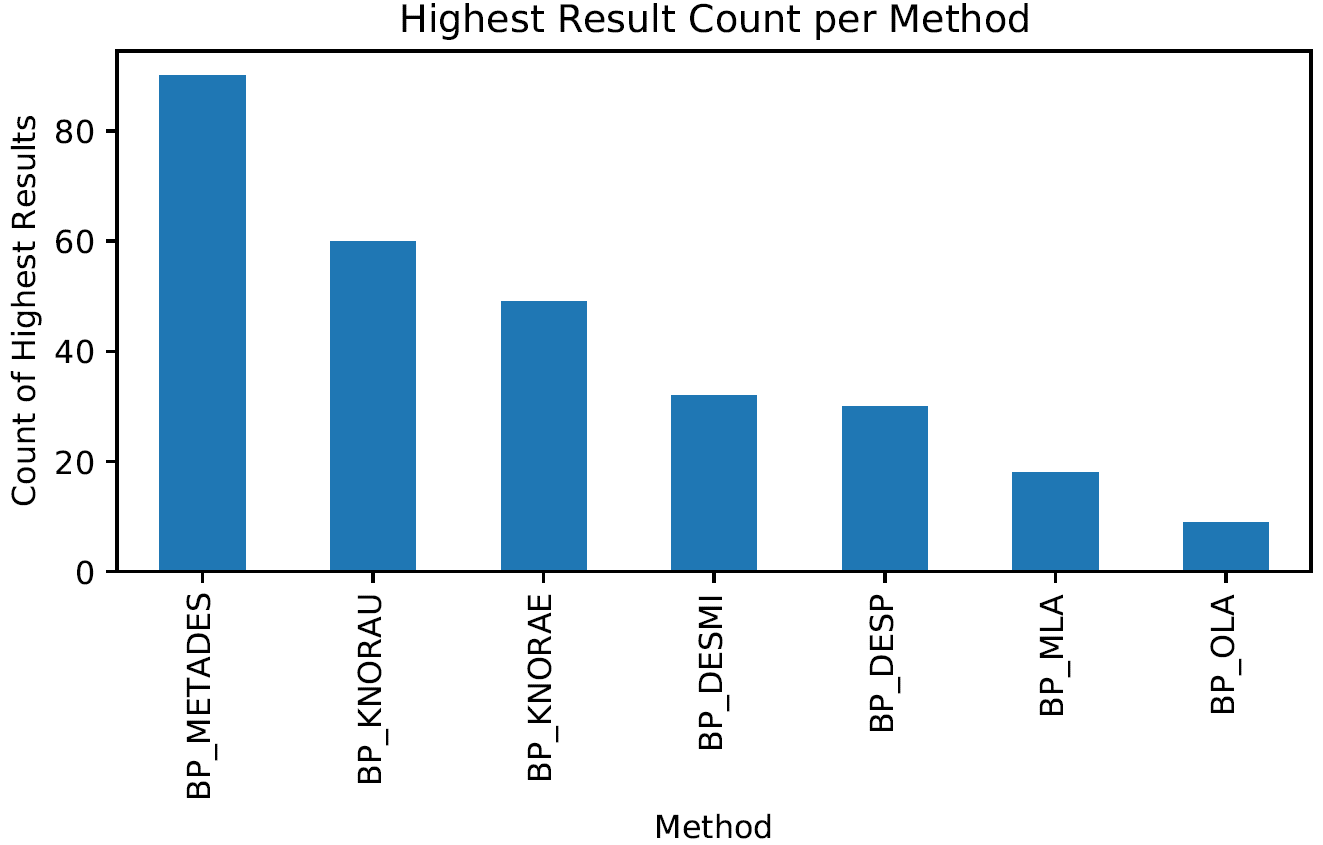}}
  \caption{Number of occurrences where each configuration attained the best result. a) Best pool generation schemes for the fixed META-DES technique. b) Best DS method for the fixed BP pool generation scheme.}
  \label{fig:bestconfig}
    \vspace{-1em}
\end{figure}

When considering META-DES as the DS algorithm, the pool generation scheme recommended by MLRS-P was the top performer in 207 out of 288 datasets, accounting for 71.87\%, while for the KNORA-E method, MLRS-P recommends the optimal pool for 228 datasets (79.15\%). In contrast, the majority baseline, which consistently uses the RF model, was only optimal for 90 datasets (31.25\%). The distribution of the best pool generation scheme for the META-DES dataset is presented in Figure~\ref{fig:bestconfig} a)~\footnote{Figures showing the best pool generation method distribution for other DS algorithms are available in the project's GitHub repository as supplementary material}.

Our MLRS-P's performance in recommending suitable classifier pools for specific DS methods significantly surpasses the baselines. Moreover, the result is also much higher than the random prediction, which corresponds to 1/7 in this multi-class classification context. This confirms MLRS-P's ability to effectively model the relationship between meta-features and pool generation schemes for DS methods. Importantly, the optimal pool generation scheme varies notably with the DS model used. For instance, while RF achieved the highest number of wins with META-DES, other DS methods like OLA and MLA found the most success with BP (Bagging with Perceptron), and KNORA-E and KNORA-U with BDT (Bagging with Decision Trees). This highlights the pivotal role of the pool generation scheme in dynamic selection and that their choice must be taken into account based on the dataset characteristics and the DS method employed.

\begin{table}
\centering
\caption{Comparison of pool recommendation accuracy between MLRS-P, majority-based, and average combination methods across different DS algorithms. Values represent the accuracy, and the number in parentheses indicates the total datasets where each method successfully recommends the optimal pool generation scheme.}
\label{result1}
\resizebox{0.40\textwidth}{!}{%
\begin{tabular}{|l|l|l|l|}
\hline
DS Method & MLRS-P & Majority & Average \\
\hline
KNORA-E & \textbf{79.16 (228)} & 31.25 (90) & 21.92 (43.00) \\
META-DES & \textbf{71.87 (207)} & 31.25 (90) & 24.84 (44.86) \\
KNORA-U & \textbf{70.13 (202)} & 36.11 (104) & 25.04 (48.29) \\
DES-MI & \textbf{78.12 (225)} & 38.19 (110) & 22.26 (42.86) \\
DES-P & \textbf{71.52 (206)} & 35.41 (102) & 25.24 (45.29) \\
MLA & \textbf{66.66 (192)} & 26.38 (76) & 24.65 (48.14) \\
OLA & \textbf{62.84 (181)} & 48.26 (139) & 20.83 (42.71) \\  
\hline
\end{tabular}%
}
\end{table}

\subsection{Scenario II: meta-learning for recommending the best DS model}

As the second type of recommendation evaluated in this work, MLRS-DS suggests a DS algorithm while a pool generation scheme is fixed. This is a critical use case for applications where a user already has a pre-trained pool of classifiers and wants to select the one more likely to maximize accuracy $(DS_o)$ from the set of possible DS models. 

Table \ref{result2} presents a comparison between the performance of MLRS-DS against the majority selection and average baselines. Each row in the table corresponds to a fixed pool generation scheme that the recommendation algorithm takes as input to recommend the most appropriate DS model. It can be seen that MLRS-DS generally obtains much higher accuracy than the Majority and Average baselines. Taking BP as a fixed pool generation scheme model, for example, we observe that MLRS-DS can successfully predict the optimal DS method in 182 datasets, corresponding to 63.19\% accuracy. The distribution of DS algorithms for the BP pool dataset is presented in Figure~\ref{fig:bestconfig} b). 

A similar behavior occurs when the recommendation system is analyzed using different fixed pool generation schemes, such as LIT (best recommendation in 207 datasets). In contrast, the accuracy of the majority recommender is 30.55\% (88 datasets) and 31.25\% (90 datasets) for LIT and BP, respectively. Thus, the results help us further confirm the hypothesis that some pools of classifiers are better suited for different DS models, and we cannot simply rely on always using a fixed top-performing DS model. 

Another observation is that although the recommendation performance is higher than the majority, average, and random baselines, the overall MLRS-DS accuracy is lower than Scenario I, indicating that recommending the best DS model is more challenging than recommending the pool.

\begin{table}
\centering
\caption{Comparison of MLRS-DS performance against the majority method and average baseline across 288 datasets. Values represent the accuracy, and the number in parentheses indicates the total datasets where each method successfully recommends the optimal DS algorithm.}
\label{result2}
\resizebox{0.4\textwidth}{!}{%
\begin{tabular}{|l|l|l|l|}
\hline
Pool Gen. & MLRS-DS & Majority & Average \\
\hline
LIT   &  \textbf{71.87 (207)} &  30.55 (88)    &    21.87 (42.71)   \\
BP    &  \textbf{63.19 (182)} &  31.25 (90)    &    30.25 (66.57)   \\
BDT   &  \textbf{62.50 (180)} &  59.02 (170)    &    26.98 (49.14)   \\
BSDT  &  \textbf{61.11 (176)} &  57.29 (165)    &    22.17 (45.14)   \\
BSP   &  \textbf{63.88 (184)} &  33.68 (97)    &    20.83 (43.00)   \\
RF    &  \textbf{59.37 (171)} &  50.00 (144)   &    27.03 (52.85)   \\
FLT   &  \textbf{57.63 (166)} &  31.25 (90)   &    23.90 (43.28)   \\  
\hline
\end{tabular}%
}
\end{table}

\subsection{Scenario III: meta-learning for recommending the pool and DS algorithm}

Scenario III, called MLRS-PDS, performs a chained recommendation. It first recommends the more suitable pool generation scheme, $C_o$, according to the meta-feature, then recommends the DS method, $DS_o$, conditional to the first choice. Thus, it consists of a multi-label prediction that recommends the whole DS pipeline in an end-to-end fashion.

Table \ref{comparison} showcases a comparison of MLRS-PDS against MLRS-P, MLRS-DS, as well as the top-4 performing pairs of pool and DS models (i.e., the ones with the highest amount of top results across the 288 datasets). Notably, in the case of MLRS-P for this analysis, the selection was based on the fixed META-DES, which was identified as the most effective dynamic selection scheme across all datasets. Similarly, for MLRS-DS, the RF model was chosen as the fixed pool generation scheme since it presented the highest number of wins among other methods. The results of each possible configuration (7 pool generation schemes $\times$ 7 DS models) per dataset can be found as supplementary material on the project's GitHub page. 

\begin{table}
\centering
\caption{The comparison between the three versions of MLRS and the baselines among 288 datasets. MLRS-P with META-DES recommends pools with a fixed META-DES DS method. MLRS-DS with RF recommends DS methods with RF as the fixed pool scheme. The four combinations below the horizontal line correspond to the top 4 (Pool, DS) configurations among the 49 possible ones.}
\label{comparison}
\small
\begin{tabular}{|l|l|}
\hline
Algorithm                          & Accuracy (wins) \\
\hline
MLRS-PDS                           & \textbf{64.93 (187)}  \\      
MLRS-P with META-DES                 & 10.06 (29)  \\
MLRS-DS with RF                      & 27.08 (78)  \\
\hline
(RF, META-DES)                     & 21.52 (62)     \\
(BP, DES-MI)                       & 11.80 (34)      \\
(BP, META-DES)                     & 10.06 (29)    \\
(BSDT, KNORA-U)                    & 4.51 (13)      \\
\hline
\end{tabular}
\end{table}

Several conclusions can be drawn from these analyses. First, when seeking the optimal solution, opting for the multi-label formulation (MLRS-PDS) and predicting both pipeline stages is advantageous. The results indicate that, for a significant number of datasets, either fixing the pool and letting the MLRS-DS suggest the DS method or fixing the DS and letting MLRS-P recommend the optimal pool generation scheme leads to a sub-optimal outcome, as it considerably limits the search space to just one decision step. Therefore, modeling the entire process through meta-learning is essential to maximize performance. Second, relying on robust, pre-existing pairs proves to be insufficient. The configuration that obtains the overall best results among the 49 possible ones (RF, META-DES) is the optimal choice for just 62 out of 288 datasets (21.52\% of the total). This analysis demonstrates that the pool generation scheme choices and their relationship with the problem characteristics and the DS model employed should not be overlooked.

\begin{figure}
  \centering
   \subfigure{\includegraphics[width=0.35\textwidth]{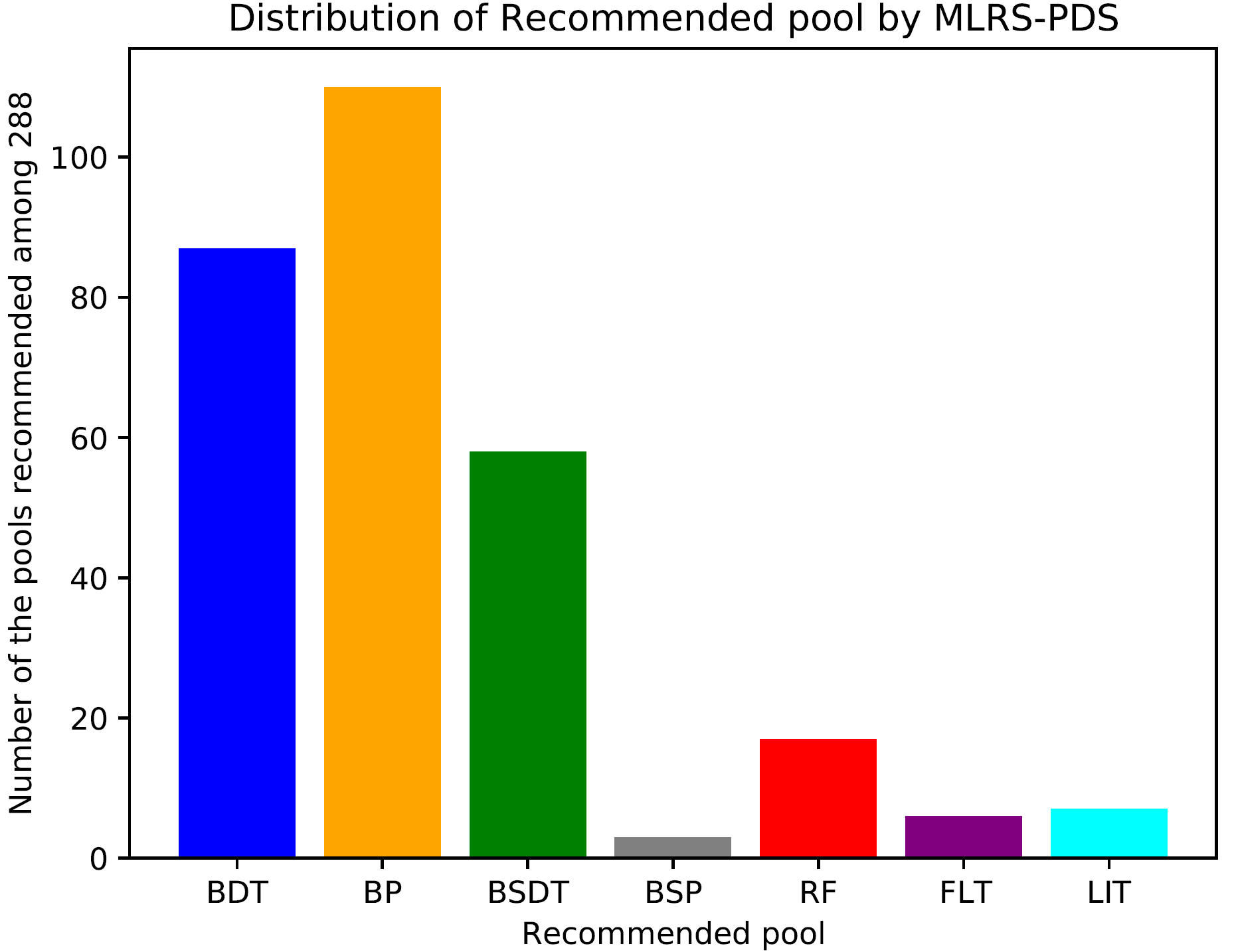}}
  \subfigure{\includegraphics[width=0.35\textwidth]{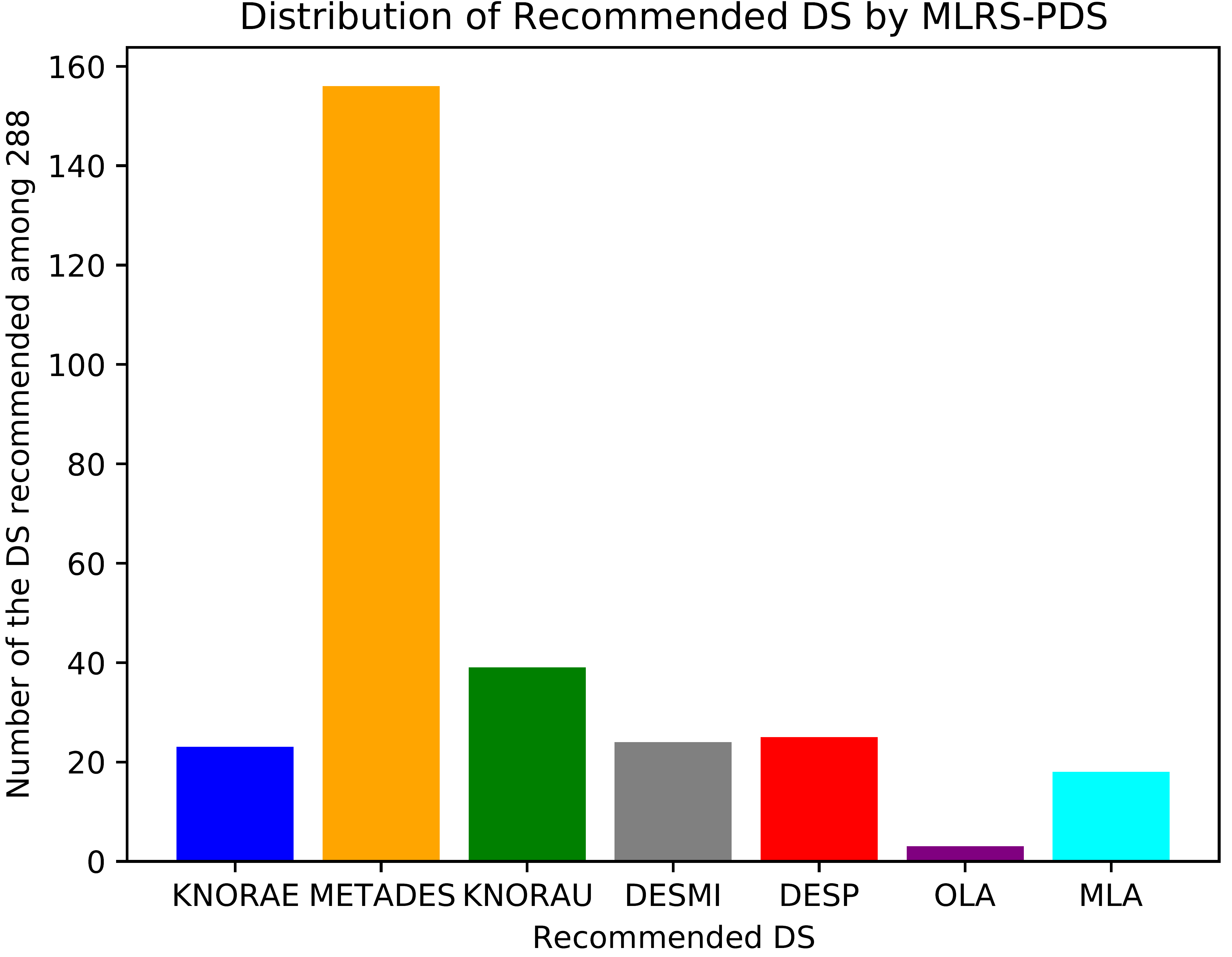}}
  \caption{Number of occurrences that each configuration attains the best result.}
  \label{fig:selectionDistribution}
  \vspace{-1em}

\end{figure}
Therefore, researchers and practitioners must avoid relying on a single, predefined pool generation scheme when comparing various DS models. Different models are based on different local assumptions and require distinct pools of classifiers to achieve optimal performance. Moreover, our proposed MLRS-PDS can be an interesting tool for practitioners who want to obtain the best performance possible using DS methods without needing to evaluate all possible configurations.

\subsection{Recommendation analysis}

Figure~\ref{fig:selectionDistribution} shows the frequency of recommended methods by the MLRS-PDS model. We can see that although some techniques are more often recommended, such as RF as the pool generation scheme and META-DES as the best DS model, there is a diversity in the recommendation scheme, demonstrating that the proposed MLRS-PDS model indeed changes its recommendation according to the problem's characteristics instead of relying on fixed robust configurations such as (RF, META-DES) that works well across the majority of cases.

\section{Conclusion}
\label{sec:conclusion}

This paper investigates two crucial yet often overlooked questions in DS research: 1) the impact of the pool generation scheme on a DS pipeline's performance, considering dataset characteristics and the DS method; and 2) a methodology for determining an optimal DS pipeline for specific datasets. In response, we present a meta-learning recommendation system that enhances dynamic selection (DS) implementation by advising on its essential design steps, namely pool generation and DS algorithm, tailored to each dataset's unique features. We developed a meta-model based on dataset-specific meta-features and a meta-target that indicates the optimal DES algorithm's performance. We propose and analyze distinct recommendation modes for user convenience: 1) MLRS-P, which recommends the best pool generation scheme when the user prefers a specific DS method, thereby optimizing its performance; 2) MLRS-DS, which suggests the most suitable DS method for a predefined pool, which is ideal for users with an existing pool model seeking the best DS technique. 3) MLRS-PDS automatically selects the optimal pair of pool and DS methods based solely on the meta-features, streamlining the design process by eliminating manual decision-making.

Our extensive empirical study on 288 diverse datasets demonstrates that MLRS recommends the correct algorithm for the three evaluated scenarios with much higher prediction performance than the usual baselines. This study demonstrates the importance of aligning classifier pools with each dataset's unique characteristics and the corresponding DS method, thereby highlighting the limitations of a one-size-fits-all strategy. Additionally, the results reveal that pool generation and its synergy with the DS method employed must not be neglected. As such, practitioners in the field must consider these conclusions when conducting further development and comparison of DS algorithms and when applying DS solutions to solve real-world problems. 

Additionally, this study highlights the effectiveness of meta-learning in developing machine learning solutions, particularly in scenarios with significant interdependence between components, where the multi-label recommendation through chained prediction can model the relationship between each design step. Future work will focus on enhancing the meta-learning framework by including recommendations for the hyperparameters of the DS models and pool generation schemes. We also aim to explore other alternatives for the meta-feature extraction process.

\bibliographystyle{IEEEtran}

\bibliography{refs}

\end{document}